\documentclass[conference]{IEEEtran}
\IEEEoverridecommandlockouts
\usepackage{cite}
\usepackage{amsmath,amssymb,amsfonts}
\usepackage{algorithmic}
\usepackage{graphicx}
\usepackage{textcomp}
\usepackage{xcolor}
\def\BibTeX{{\rm B\kern-.05em{\sc i\kern-.025em b}\kern-.08em
    T\kern-.1667em\lower.7ex\hbox{E}\kern-.125emX}}
\begin{document}

\title{Thermal Image-based Fault Diagnosis in Induction Machines
via Self-Organized Operational Neural Networks}

\author{
\IEEEauthorblockN{1\textsuperscript{st} Sertac Kilickaya}
\IEEEauthorblockA{\textit{Department of Computing}\\
\textit{Sciences}\\
\textit{Tampere University}\\
Tampere, Finland \\
sertac.kilickaya@tuni.fi}
\and
\IEEEauthorblockN{2\textsuperscript{nd} Cansu Celebioglu}
\IEEEauthorblockA{\textit{Department of Information } \\
\textit{Engineering}\\
\textit{University of Padova}\\
Padova, Italy \\
cansu.celebioglu@studenti.unipd.it}
\and
\IEEEauthorblockN{3\textsuperscript{rd} Levent Eren}
\IEEEauthorblockA{\textit{Department of Electrical } \\
\textit{and Electronics Engineering}\\
\textit{Izmir University of Economics}\\
Izmir, Turkey \\
levent.eren@ieu.edu.tr}
\and
\IEEEauthorblockN{4\textsuperscript{th} Murat Askar}
\IEEEauthorblockA{\textit{Department of Electrical } \\
\textit{and Electronics Engineering}\\
\textit{Izmir University of Economics}\\
Izmir, Turkey \\
murat.askar@ieu.edu.tr}
}
\maketitle

\begin{abstract}
Condition monitoring of induction machines is crucial to prevent costly interruptions and equipment failure. Mechanical faults such as misalignment and rotor issues are among the most common problems encountered in industrial environments. To effectively monitor and detect these faults, a variety of sensors, including accelerometers, current sensors, temperature sensors, and microphones, are employed in the field. As a non-contact alternative, thermal imaging offers a powerful monitoring solution by capturing temperature variations in machines with thermal cameras. In this study, we propose using 2-dimensional Self-Organized Operational Neural Networks (Self-ONNs) to diagnose misalignment and broken rotor faults from thermal images of squirrel-cage induction motors. We evaluate our approach by benchmarking its performance against widely used Convolutional Neural Networks (CNNs), including ResNet, EfficientNet, PP-LCNet, SEMNASNet, and MixNet, using a Workswell InfraRed Camera (WIC). Our results demonstrate that Self-ONNs, with their non-linear neurons and self-organizing capability, achieve diagnostic performance comparable to more complex CNN models while utilizing a shallower architecture with just three operational layers. Its streamlined architecture ensures high performance and is well-suited for deployment on edge devices, enabling its use also in more complex multi-function and/or multi-device monitoring systems.
\end{abstract}


\section{Introduction}
Induction machines are widely used in applications such as heating, ventilation, and air conditioning (HVAC) systems, pumps, and conveyor belts across industrial, commercial, and residential sectors. Their widespread use is attributed to their efficiency, durability, reliability, and cost-effectiveness.
Despite their reliability, induction machines are not completely immune to failure. 
Repeated and long-term use of induction machines can contribute to the deformation and wear of parts, which decreases the operational life of asynchronous machines and affects their performance \cite{bib1_1}. Induction machine faults are categorized into two distinct classes: electrical and mechanical. Mechanical breakdowns can occur due to issues such as bearing failures and rotor imbalances, whereas electrical faults can arise from overloads, stator open phases, and short circuits. These faults are critical for induction machines and can significantly impact their performance, resulting in reduced efficiency, unexpected shutdowns, and potentially expensive repairs \cite{bib1}. 

Timely detection and diagnosis of induction machine failures are essential to maintain operational efficiency and prevent catastrophic breakdowns. In the literature, a range of sensors, including accelerometers, current sensors, temperature sensors, and microphones, are commonly used for fault diagnosis. For example, \cite{bib3} utilizes Vibration Spectrum Imaging (VSI) to convert normalized spectral amplitudes from segmented vibration signals into images, which are then used to train a convolutional neural network (CNN) for bearing fault classification. The study in \cite{bib4} presents a smartphone-based system for diagnosing faults in rotating machines by leveraging the device's audio recording capability to capture machine sounds. The collected audio data is used to train a 1D CNN, which is then integrated into a mobile application for real-time and cost-effective fault diagnosis.  

\begin{figure*}
    \centering
    \includegraphics[width=1\linewidth]{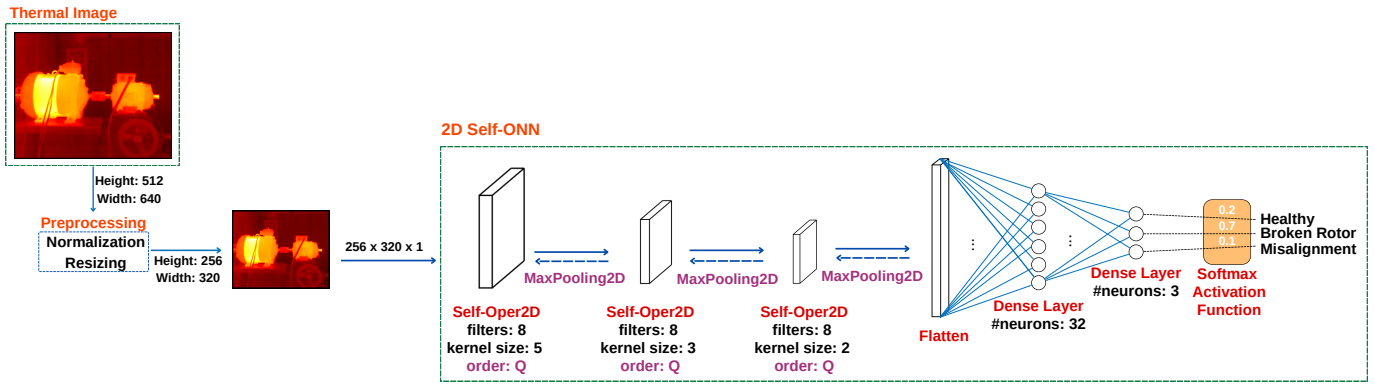}
   \caption{Proposed 2D Self-ONN architecture with preprocessing steps for fault diagnosis using thermal images from \cite{bib13}.}
   \label{fig:overview}
\end{figure*}

Beyond these approaches, image processing techniques that utilize thermal images have also been proposed as valuable tools for fault detection. In particular, physical changes on the motor's surface, and visual anomalies such as rotor and stator defects can be effectively diagnosed with thermal cameras or high-resolution imaging devices. Thermal cameras can detect abnormal temperature changes that occur before or after mechanical or electrical defects using infrared rays. This approach allows for non-invasive machine health monitoring and enables the detection of overheated components and other thermal anomalies. 

Many studies in recent years have used thermal imaging to diagnose faults in induction machines and their external components. For instance, \cite{bib5} introduced a method for extracting features from thermal images called Method of Areas Selection of Image Differences (MoASoID). By analyzing different training sets to identify areas with the most significant changes, these regions were utilized for recognition. Feature vectors were extracted using the MoASoID and image histograms, while classification was performed using techniques such as Nearest Neighbour (NN), K-means, and Backpropagation Neural Network (BNN). Another approach was introduced using the Hue, Saturation, and Value (HSV) color model instead of the grayscale model \cite{bib6}. Five image segmentation methods, namely Sobel, Prewitt, Roberts, Canny, and Otsu, were then employed to segment the Hue region, and key statistical features were extracted to differentiate between motor faults under diverse load conditions. \cite{bib7} proposed a technique for analyzing thermal images of commutator motors (CMs) and single-phase induction motors (SIMs) for fault diagnosis using novel feature extraction methods, such as Differences of Arithmetic Mean with Otsu’s Method (DAMOM) and Differences of Arithmetic Mean with 20 Highest Peaks (DAM20HP), classified through NN and Long Short-term Memory (LSTM) algorithms. \cite{bib8} employed a method to classify thermal images of rotating machinery, using features extracted by a CNN for fault diagnosis with a Softmax regression classifier. \cite{bib9} employed thermography and the InceptionV3 model, utilizing Contrast Limited Adaptive Histogram Equalization (CLAHE) for image preprocessing and a Squeeze-and-Excitation (SE) channel attention mechanism. \cite{bib10} suggested a technique for rotor-bearing frameworks using a two-stage parameter transfer approximation and infrared thermal images. It utilizes a scaled exponential linear unit (SELU) and modified stochastic gradient descent (MSGD) to build an enhanced convolutional neural network (ECNN), which was primarily trained with a stacked convolutional auto-encoder (CAE) on unlabeled source-domain thermal images. Model parameters were then transferred to fit the ECNN for a target-domain, with performance evaluated on thermal images collected under different speeds. Furthermore, \cite{bib12} analyzed thermal images of a three-phase induction motor for four distinct bearing conditions: outer race defect, inner race defect, lack of lubrication, and healthy bearing. They preprocess thermal images using Discrete Wavelet Transform (DWT), extract and choose features with Mahalanobis distance, and classify the bearing conditions using a Support Vector Machine (SVM) classifier.

Regardless of the type of data, whether sound, vibration, or image, these approaches can be computationally intensive and require significant processing power and time, which may limit their performance in real-time systems. To enhance both performance and computational efficiency in machine fault diagnosis, 1D CNNs have been utilized in several studies \cite{cnn_1},\cite{cnn_2}. To further the improve performance of CNNs while adding minimal computational overhead, Self-Organized Operational Neural Networks (Self-ONNs) were proposed in \cite{kiranyaz2021self}. These networks enhance the performance of CNNs through their non-linear neuron model and high heterogeneity in its layers. For instance, in \cite{rotor_onn}, Self-ONNs were used to enhance the detection and classification of broken rotor bars, a type of malfunction that is common in induction motors and remarkably affects their lifespan. The study conducted in \cite{bib2} explored the implementation of domain-adaptive fault diagnosis using Self-ONNs and vibration data. It compares the performance of 1D CNNs and Self-ONNs, ultimately proposing the use of Self-ONNs as feature extractors within domain-adversarial neural networks (DANN) to enhance adaptation performance across varying loading conditions.

\begin{figure*}[htbp]
\centerline{\includegraphics[width=0.8\linewidth]{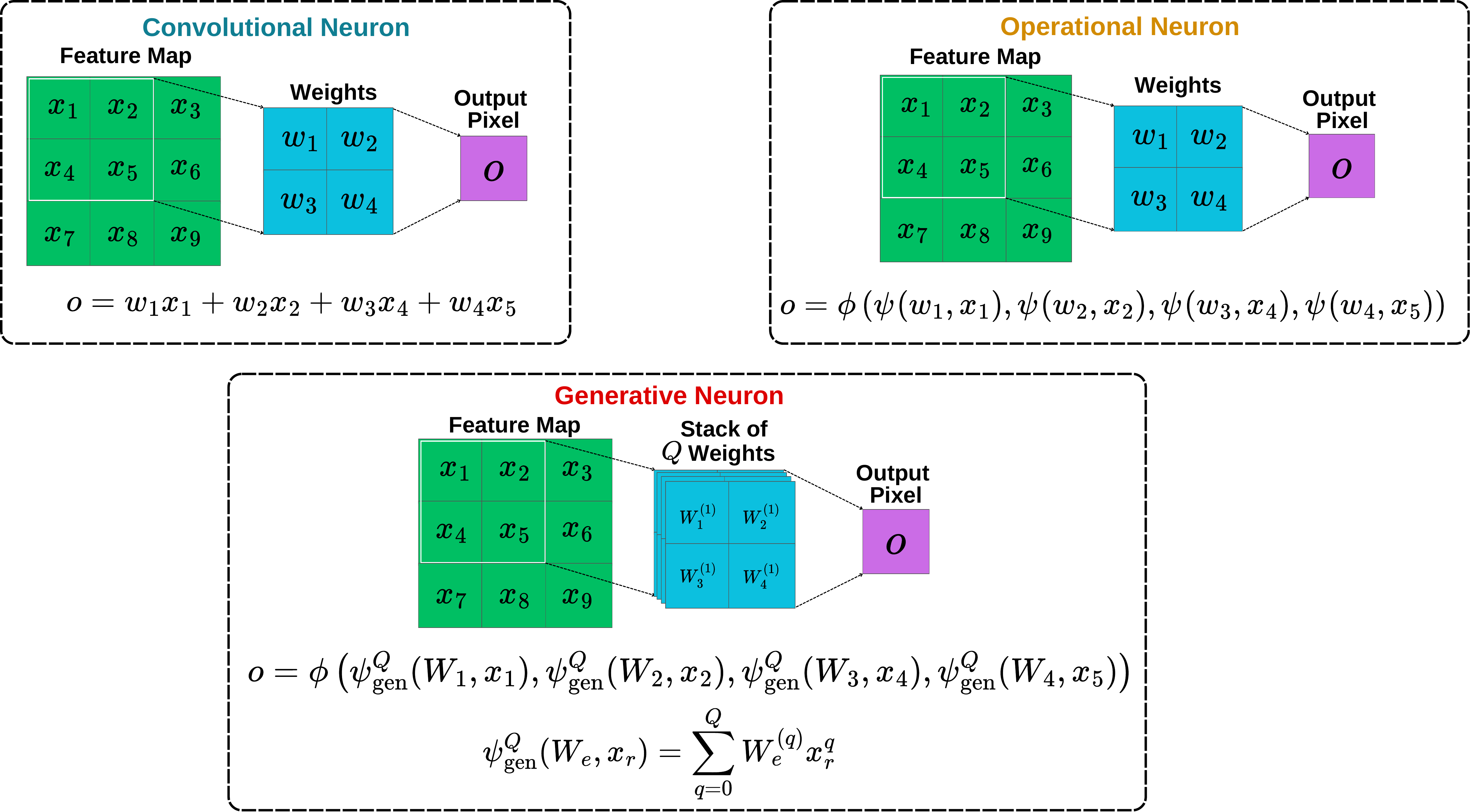}}
\caption{Visual representation of the formulations for convolutional, operational, and self-organizing (generative) neurons \cite{malik2021self}.}
\label{fig1}
\end{figure*} 

Our study uses a thermal image dataset \cite{bib13} of squirrel cage induction motors, including images of both healthy motors and those with misalignment and broken rotor faults. We evaluate the proposed 2D Self-ONN models and compare their performance against several established CNN architectures referenced in \cite{bib13}. The performances of both the baseline CNN models and the proposed 2D Self-ONN model are evaluated using metrics such as accuracy, recall, precision, and F1-score. Additionally, we compute the number of trainable parameters and inference duration for each model, making comparisons regarding their computational efficiency and complexity. Our research demonstrates that 2D Self-ONNs can achieve comparable, and in some cases superior, fault diagnosis accuracy while utilizing significantly fewer layers and trainable parameters than traditional deep CNN models. Fig. \ref{fig:overview} depicts an overview of the proposed 2D Self-ONN model with preprocessing steps.

The contents of the paper are organized as follows: Section II provides an overview of the theoretical foundation underlying the 2D Self-ONN model. Section III outlines the experimental setup, describes the dataset, and details the preprocessing steps employed. Section IV presents the fault diagnosis performance results of the proposed 2D Self-ONN models and baseline CNNs on the thermal image dataset. It also includes an analysis of the number of trainable parameters and inference times associated with each method. The paper concludes with a summary of the findings and recommendations for future work.

\section{Self-organized Operational Neural Networks}
In this work, we employ 2D Self-Organized Operational Neural Networks (Self-ONNs) for fault diagnosis using thermal images. This section aims to explain the concept of generative neurons and their integration into 2D Self-ONNs. 

In a 2D convolutional layer, given the output of layer \((l-1)\), the pre-activation output of the \(k\)-th convolutional neuron in layer \(l\) can be computed as:

\begin{equation}
x_{k}^{l}(m,n)=\sum_{r=0}^{K-1}\sum_{t=0}^{K-1}w_{k}^{l}(r,t)y^{l-1}(m+r,n+t), \label{eq1}
\end{equation}

where ${w}_{k}^{l} \in \mathbb{R}^{K \times K}$ is the weight kernel, ${x}_{k}^{l} \in \mathbb{R}^{M \times N}$ is the input feature map, and ${y}^{l-1} \in \mathbb{R}^{M \times N}$ is the output of $(l-1)$-th layer.

In Operational Neural Networks (ONNs) \cite{kiranyaz2020operational}, the operational neuron generalizes the standard convolution operation as follows:

\begin{equation}
\overline{x_{k}^{l}}(m,n)=\phi_{k}^{l}(\psi_{k}^{l}(w_{k}^{l}(r,t),y^{l-1}(m+r,n+t)))_{(r,t)=(0,0)}^{(K-1,K-1)} 
\label{eq2}
\end{equation}

where $\psi_{k}^{l}(\cdot): \mathbb{R}^{MN \times K^2} \to \mathbb{R}^{MN \times K^2}$ and $\phi_{k}^{l}(\cdot): \mathbb{R}^{MN \times K^2} \to \mathbb{R}^{MN}$ represent the nodal and pool operators, respectively, associated with the $k$-th neuron of the $l$-th layer. It is evident that the convolutional neuron represents a special case of an operational neuron, defined by nodal operator $\psi(x,y) = x*y$ and pooling function $\phi(\cdot) = \sum_{j}$.

In ONNs, the Greedy Iterative Search (GIS) algorithm is commonly used to iteratively examine a range of potential operators, aiming to identify the optimal combination of nodal and pool operators. Once the best operators are determined, they are assigned to all neurons within the corresponding hidden layer, resulting in the final configuration of the ONN. However, the traditional ONN architecture has several limitations \cite{kiranyaz2021self}. It restricts heterogeneity by applying the same operator set to all neurons within a layer. In addition, a significant challenge lies in selecting suitable candidate nodal and pool operators prior to network training, along with the associated computational overhead involved in this process. This approach can introduce bias, as the initial selection of operators may influence the network's overall performance. Self-ONNs were proposed to address these challenges by incorporating a generative neuron model \cite{kiranyaz2021self}.

The nodal operators within a self-organized operational layer are derived through the application of Taylor series function approximation to create optimal non-linear functions. The nodal transformation in a generative neuron can be expressed in the following general form:

\begin{equation}
\begin{split}
\widetilde{\psi_{k}^{l}}&(w_{k}^{l(Q)}(r,t),y^{l-1}(m+r,n+t))
\\&=\sum_{q=1}^{Q}w_{k}^{l(Q)}(r,t,q)(y^{l-1}(m+r,n+t))^{q} 
\end{split}
\label{eq3}
\end{equation}

In Equation (\ref{eq3}), $Q$ serves as a hyperparameter that controls the degree of the Taylor polynomial approximation, thus it affects the level of non-linearity. Additionally, the weights $w_{k}^{l(Q)}$ now include $Q$ times the number of learnable parameters found in the convolutional model. As a result, $w_{k}^{l} \in \mathbb{R}^{K \times K}$ is replaced by $w_{k}^{l(Q)} \in \mathbb{R}^{K \times K \times Q}$. During training, the weights $w_{k}^{l(Q)}$ are updated with the standard back-propagation (BP) algorithm, resulting in non-linear transformations \cite{kiranyaz2021self}. 

By defining the pooling operator as a summation function, we can represent the self-organized operational layer through the convolutional model, and find the output of the generative neuron as follows:

\begin{equation}
\widetilde{x_{k}}^{l}(m,n)=\sum_{r=0}^{K-1}\sum_{t=0}^{K-1}\sum_{q=1}^{Q}w_{k}^{l(Q)}(r,t,q)(y^{l-1}(m+r,n+t))^{q} 
\label{eq4}
\end{equation}

In simpler terms, it might be expressed as:

\begin{equation}
\widetilde{x_{k}}^{l}=\sum_{q=1}^{Q}\operatorname{Conv}2D(w_{k}^{l(Q)},(y^{l-1})^{q}) 
\label{eq5}
\end{equation}

Therefore, the formulation can be realized through the application of $Q$ 2D convolution operations, where for $Q=1$, it reduces to the standard 2D convolution. To better illustrate the concepts, the formulations for 2D convolutional, operational, and self-organized (generative) neurons are shown in Fig. \ref{fig1} \cite{malik2021self}.

\section{Dataset and Preprocessing}

\subsection{Experimental Setup and Dataset}\label{AA}

\begin{figure}[htbp]
\centerline{\includegraphics[width=0.7\linewidth]{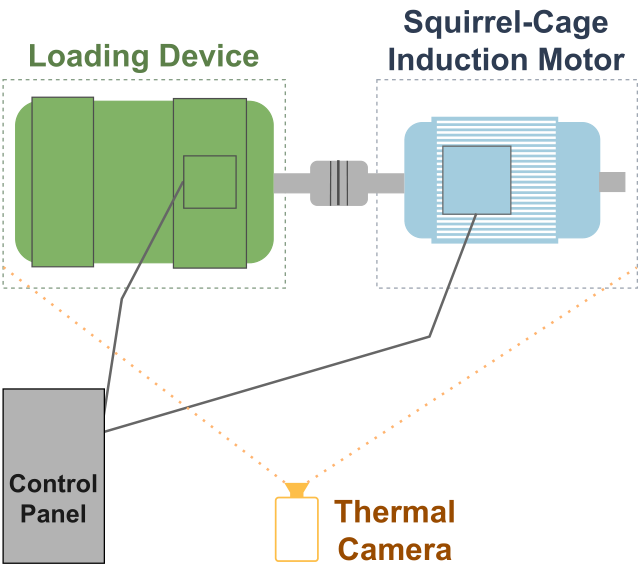}}
\caption{Experimental setup \cite{bib13}.}
\label{setup}
\end{figure} 

The experiments were conducted using a 3-phase squirrel cage induction motor, specifically the Celma Indukta Sh 90L-4 model. This motor has a rated power of $1.5$ kW, and runs at a speed of $1410$ rpm. The experimental setup is depicted in Fig. \ref{setup} \cite{bib13}. Based on the experiments conducted with squirrel cage induction motors exhibiting two different fault types, three distinct classes emerged: healthy, misalignment, and broken rotor. In the healthy class, the induction motor operates normally under standard conditions. The misalignment class represents instances where a misalignment occurred between the motor shaft and loading device, resulting in a loss of parallelism. Meanwhile, the broken rotor class features a squirrel-cage rotor with damaged cages. The experimental study examined four variations of misalignment by progressively increasing the shift between the shafts, along with three scenarios for the broken rotor case with 1, 3, and 6 broken rotor bars. In all configurations, current loads were varied from 0 A (no-load) to 6 A, increasing in steps of 2 A \cite{bib13}.

As shown in Fig. \ref{setup}, the setup was monitored using a WIC camera, manufactured by Workswell \cite{cam_wic}. This camera features a resolution of $640$×$512$ pixels and can measure temperatures ranging from -$40$°C to +$550$°C, with a thermal sensitivity of $0.03$°C. It captures images at a frame rate of $4$ frames per second (FPS). Given this frame rate, approximately $120$ images were obtained during each experiment through continuous image collection over a 30-second period.

\begin{table}[htbp]
\caption{Dataset distribution for 5-fold cross-validation of the proposed method}
\begin{center}
\begin{tabular}{|c|c|c|c|c|c|c|}

\hline
\textbf{ } & \multicolumn{6}{|c|}{\textbf{Fold}}\\
\hline
\textbf{Class} & 1 & 2 & 3 & 4 & 5 & \textbf{Total} \\
\hline
Healthy & $449$ & $449$ & $449$ & $449$ & $448$ & $2244$ \\
\hline
Misalignment & $360$ & $360$ & $360$ & $359$ & $360$ & $1799$ \\
\hline
Broken Rotor & $322$ & $322$ & $322$ & $322$ & $322$ & $1610$ \\
\hline

\end{tabular}
\label{fold_distribution}
\end{center}
\end{table}

The experiments were conducted using two different types of couplings between the motor and loading device. For coupling 1, thermal images are available for healthy motor condition and misalignment. In contrast, coupling 2 includes data for healthy motor, misalignment, and broken rotor faults. The entire dataset can be categorized into three distinct classes: healthy, misalignment, and broken rotor. In total, it contains $5,653$ frames, comprising $2,244$ images for the healthy class, $1,799$ for misalignment, and $1,610$ for broken rotor.

\begin{table}[htbp]
\caption{Temperature ranges recorded for individual experiments \cite{bib13}}
\begin{center}
\begin{tabular}{|c|c|c|c|c|}
\hline
\textbf{} & \textbf{Min} & \textbf{Max} & \textbf{Mean} & \textbf{Standart} \\ 
\textbf{Class} & \textbf{Value} & \textbf{Value} & \textbf{Value} & \textbf{Deviation} \\
\hline
Healthy & $23.00$°C & $82.43$°C & $38.62$°C & $8.86$°C  \\
\hline
Misalignment & $25.52$°C & $104.99$°C & $40.31$°C & $12.04$°C \\
\hline
Broken Rotor & $24.86$°C & $83.30$°C & $41.24$°C & $11.09$°C \\
\hline
\end{tabular}
\label{temperature}
\end{center}
\end{table}

In the baseline study \cite{bib13}, the dataset was randomly divided into five folds for cross-validation without maintaining class proportions. In our study, the dataset was divided into five equal folds, preserving the consistent class proportions of healthy, misalignment, and broken rotor without shuffling the data. The distribution of the dataset across the folds is presented in Table \ref{fold_distribution}, and Table \ref{temperature} presents the minimum, maximum, average, and standard deviation of the measured temperature ranges for each class in the dataset. Another challenge in the dataset is the temperature rise caused by varying current loads applied to each class.

\begin{figure}
    \centering
    \includegraphics[width=1\linewidth]{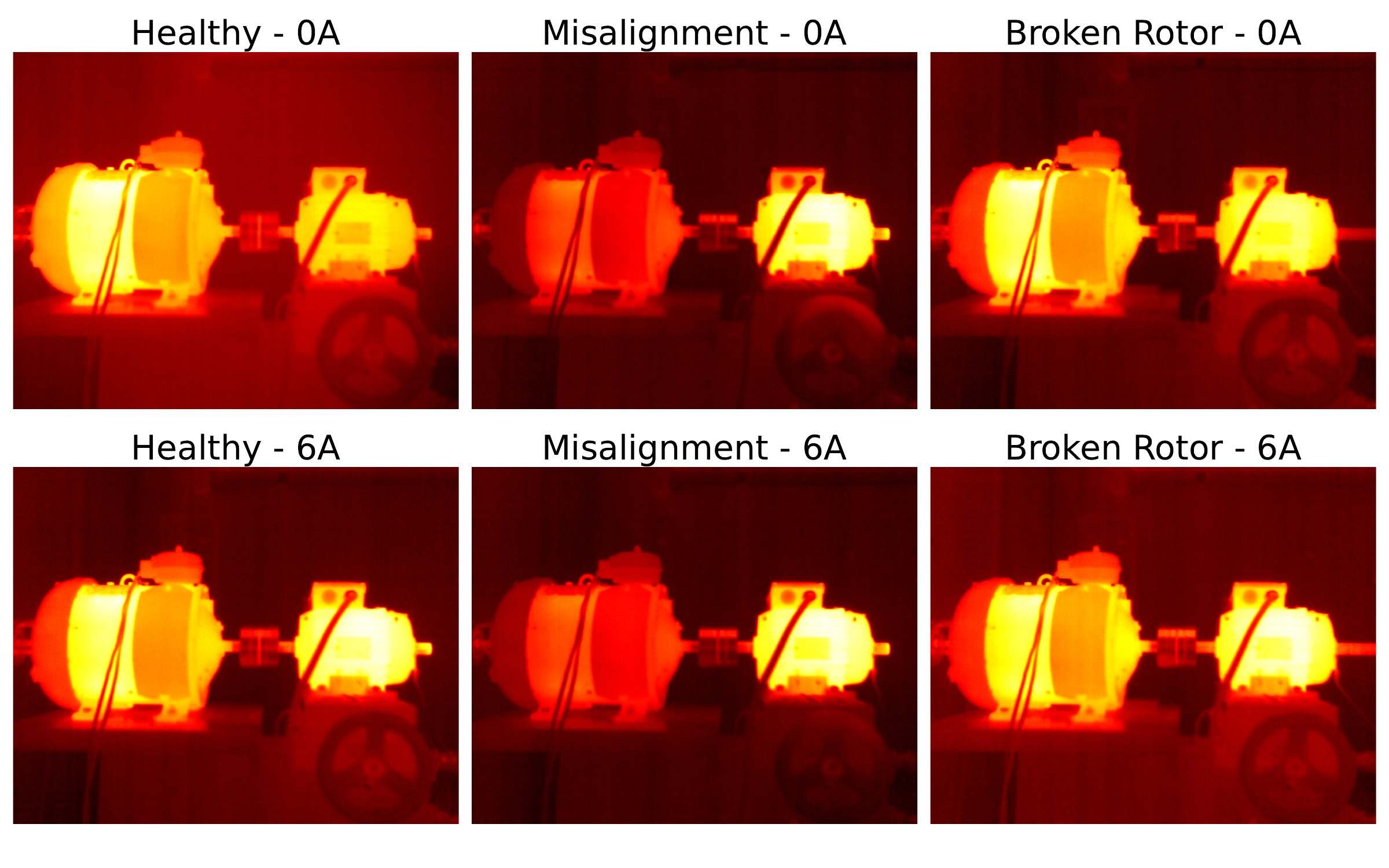}
    \caption{Examples of thermal images for a healthy motor, and motors with misalignment and broken rotor faults, under no-load and 6 A current load for coupling 2 \cite{bib13}.}
    \label{fig:thermal_6images}
\end{figure}

Fig. \ref{fig:thermal_6images} shows examples of thermal images for healthy, misaligned, and broken rotor faulty motors under no-load and 6 A current-loaded conditions. From the figure, it is evident that as the current load in the healthy motor increases, the thermal images begin to resemble those of the broken rotor faulty motor, which complicates the classification within the dataset.

\subsection{Preprocessing}\label{AA}
Thermal images are preprocessed through a two-step pipeline that includes resizing and normalization. During the resizing step, the images are reduced from their original dimensions of $640$x$512$ pixels to a target size of $320$x$256$ pixels. After resizing, pixel values are normalized to a range of $[0, 1]$. This step facilitates stable and efficient learning by aligning all pixel values to a comparable scale. The thermal images in the dataset are in $16$-bit format, with pixel values ranging from $0$ to $65,535$. The normalization step for the images can be expressed as:

\begin{equation}
 I_{\text{norm}}(i,j) = \frac{I_{\text{array}}(i,j) - \text{min}(I_{\text{array}})}{\text{max}(I_{\text{array}}) - \text{min}(I_{\text{array}}) + \epsilon}
 \label{eq6}
\end{equation}

where \( I_{\text{array}}(i,j) \) indicates the pixel value at coordinates \((i, j)\) in the image array, \( \text{min}(I_{\text{array}}) \) and \( \text{max}(I_{\text{array}}) \) are the minimum and maximum pixel values in the image, and \( \epsilon \) is a small constant added to prevent division by zero.

\begin{table*}[htbp]
\caption{Average performance metrics for benchmarked neural network models in $5$-fold cross-validation.}
\begin{center}
\begin{tabular}{|c|c|c|c|c|c|}
\hline
\multicolumn{2}{|c|}{\textbf{Model}} & \textbf{Accuracy} & \textbf{F1-Score} & \textbf{Precision} & \textbf{Recall} \\
\hline
\multicolumn{2}{|c|}{PP-LCNet 50 \cite{bib13}}& $0.924 \pm 0.078$& 
$0.939 \pm 0.066$ & 
$0.979 \pm 0.045$ &
$0.924 \pm 0.078$

 \\
\hline
\multicolumn{2}{|c|}{PP-LCNet 75 \cite{bib13}}& $0.920 \pm 0.096$&
$0.928 \pm 0.086$&
$0.954 \pm 0.072$&
$0.920 \pm 0.096$

 \\
\hline
\multicolumn{2}{|c|}{PP-LCNet 100 \cite{bib13}}& $0.880 \pm 0.102$&
$0.892 \pm 0.088$&
$0.930 \pm 0.053$&
$0.880 \pm 0.102$

 \\
\hline
\multicolumn{2}{|c|}{SEMNASNet 50 \cite{bib13}}& $0.901 \pm 0.091$&
$0.920 \pm 0.080$&
$0.964 \pm 0.075$&
$0.901 \pm 0.091$

 \\
\hline
\multicolumn{2}{|c|}{SEMNASNet 75 \cite{bib13}}& $0.913 \pm 0.054$&
$0.935 \pm 0.045$&
$0.979 \pm 0.043$&
$0.913 \pm 0.054$

 \\
\hline
\multicolumn{2}{|c|}{SEMNASNet 100 \cite{bib13}}& $0.878 \pm 0.108$&
$0.911 \pm 0.071$&
$0.978 \pm 0.043$&
$0.878 \pm 0.108$

 \\
\hline
\multicolumn{2}{|c|}{ResNet10t \cite{bib13}}& $0.831 \pm 0.167$&
$0.866 \pm 0.136$&
$0.952 \pm 0.076$&
$0.831 \pm 0.167$ 

 \\
\hline
\multicolumn{2}{|c|}{ResNet18 \cite{bib13}} & $0.850 \pm 0.172$&
$0.879 \pm 0.136$&
$0.960 \pm 0.061$&
$0.850 \pm 0.172$

 \\
\hline
\multicolumn{2}{|c|}{ResNet34 \cite{bib13}}& $0.925 \pm 0.093$&
$0.929 \pm 0.086$&
$0.957 \pm 0.061$&
$0.925 \pm 0.093$

 \\
\hline
\multicolumn{2}{|c|}{EfficientNet B0 \cite{bib13}}& $0.896 \pm 0.141$&
$0.907 \pm 0.131$&
$0.946 \pm 0.078$&
$0.896 \pm 0.141$

 \\
\hline
\multicolumn{2}{|c|}{EfficientNet B1 \cite{bib13}*}& \textbf{\boldmath{ $0.932 \pm 0.086$}}&
\textbf{\boldmath{ $0.941 \pm 0.076$ }}&
\textbf{\boldmath{ $0.969 \pm 0.046$ }}&
\textbf{\boldmath{ $0.932 \pm 0.086$ }}
 \\
\hline
\multicolumn{2}{|c|}{EfficientNet B2 \cite{bib13}}& $0.854 \pm 0.137$&
$0.864 \pm 0.129$&
$0.899 \pm 0.103$&
$0.854 \pm 0.137$

\\
\hline
\multicolumn{2}{|c|}{MixNet S \cite{bib13}}& $0.904 \pm 0.088$& 
$0.916 \pm 0.077$&
$0.947 \pm 0.050$&
$0.904 \pm 0.088$

 \\
\hline
 & \multicolumn{1}{l|}{$Q = 1$} & $0.887 \pm 0.106$&
$0.880 \pm 0.116$&
$0.931 \pm 0.058$&
$0.887 \pm 0.106$

 \\
\cline{2-6} 
 & \multicolumn{1}{l|}{$Q = 2$**} & \textbf{\boldmath{$0.942 \pm 0.070$}}&
\textbf{\boldmath{ $0.941 \pm 0.072$}} &
\textbf{\boldmath{ $0.960 \pm 0.045$}} &
\textbf{\boldmath{ $0.942 \pm 0.070$}}

 \\
\cline{2-6} 
2D Self-ONN & \multicolumn{1}{l|}{$Q = 3$} & $0.940 \pm 0.074$&
$0.938 \pm 0.076$&
$0.959 \pm 0.047$&
$0.940 \pm 0.074$

 \\
\cline{2-6} 
& \multicolumn{1}{l|}{$Q = 4$} & $0.897 \pm 0.096$&
$0.892 \pm 0.102$&
$0.935 \pm 0.055$&
$0.897 \pm 0.096$

 \\ \cline{2-6}
&\multicolumn{1}{l|} {$Q = 5$} & $0.907 \pm 0.083$&
$0.904 \pm 0.086$&
$0.934 \pm 0.052$&
$0.907 \pm 0.083$

 \\
\hline
\multicolumn{4}{l}{* The best result from the reference study.} \\
\multicolumn{4}{l}{** The best result in all our proposed methods (2D Self-ONN).} 
\end{tabular}
\label{all_metrics}
\end{center}
\end{table*}

\section{Experiments and Results}
This section presents the evaluation criteria and fault diagnosis performance results of our proposed 2D Self-ONN model, compared to the CNN architectures from the baseline study \cite{bib13}. We used metrics such as accuracy, F1-score, precision, and recall. In addition, we compare the computational efficiency by analyzing the number of trainable parameters and inference durations to ensure a comprehensive evaluation of each model's performance and limitations.

In \cite{bib13}, a non-invasive solution for near-real-time monitoring was proposed using edge devices. The study explores various well-known CNN architectures, including PP-LCNet \cite{pplcnet} for its CPU efficiency, and several ResNet models \cite{resnet10}. It also examines MNASNet \cite{MNASNet}, particularly the SEMNASNet variant that incorporates Squeeze-and-Excitation mechanisms, as well as EfficientNet variants \cite{efficientnet}, which effectively scale network dimensions to enhance performance. Additionally, the S version of MixNet is utilized for its robust feature extraction capabilities. In the reference study, the preprocessing stage involved converting thermal images from a 16-bit unsigned integer to a $32$-bit floating-point representation. During this process, min-max scaling was applied to normalize the pixel values to the range of $[0, 1]$. The minimum and maximum values for all thermal images across the dataset were determined and predefined for this min-max scaling. They partitioned the dataset into $5$ folds to conduct $5$-fold cross-validation. For each test fold, the remaining three folds were used for training, while one fold served as the validation set. The AdamW optimizer \cite{adamW} was utilized with a learning rate of $0.00003$, and cross-entropy was employed as the loss function. Depending on the CNN model, a batch size of either $8$ or $16$ was used, and each model was trained for a maximum of $300$ epochs, implementing early stopping with a patience of $10$ epochs.

In this study, we resized thermal images to $320$x$256$ pixels and applied min-max normalization using Equation \ref{eq6}. The entire dataset was divided into $5$ equal folds, maintaining consistent class proportions across each fold. For each test fold, the remaining three folds were designated for training, while one fold served as the validation set, in line with the methodology of the reference study. Adam \cite{adam} was used as the optimizer and the learning rate was initialized to $0.001$. Two callbacks were created during training. Early stopping was implemented to monitor the validation loss, stopping training if there is no improvement for $5$ consecutive epochs. Additionally, a learning rate adjustment strategy was used, which reduced the learning rate by $50$\% if the validation loss did not improve for $3$ epochs, with a minimum learning rate of $0.00005$. We restored the best weights corresponding to the minimum validation loss and evaluated the model on the test fold. The batch size was set to $16$, and the cross-entropy loss was utilized during training. 

The proposed 2D Self-ONN model, with three self-operational layers, is illustrated in Fig. \ref{fig:overview}. Here, $Q$ corresponds to the order of the Taylor polynomial approximation in 2D self-organized operational layers. For $Q=1$, the network reduces to a 2D CNN. Each self-operational layer has $8$ filters with kernel sizes of $5$x$5$, $3$x$3$, and $2$x$2$ respectively, and employs the $tanh$ activation function. All self-operational layers are followed by a max-pooling operation with a $2$x$2$ kernel and a stride of 2 to downsample the data. After these layers, the output is flattened and fed into a fully connected layer with $32$ units, employing the $tanh$ activation function. The output layer consists of 3 neurons and utilizes Softmax activation to classify the input into one of the predefined classes, generating output probabilities.

Table \ref{all_metrics} offers a detailed summary of the accuracy, F1-score, precision, and recall for the models examined in the reference paper \cite{bib13}, as well as the 2D CNN $(Q=1)$ and Self-ONN models $Q=2$ to $Q=5$ introduced in this study. These metrics assess each classifier's effectiveness in differentiating between specific cases and non-cases. While precision shows the fraction of correctly identified events to all detected events, recall indicates the percentage of correctly classified cases out of all actual cases. Lastly, F1-score identifies the harmonic mean of the model’s Recall and Precision. These metrics are calculated according to the counts of false negatives (FN), false positives (FP), true negatives (TN), and true positives (TP) as follows: 

\begin{equation}
\text{Accuracy} = \frac{TP + TN}{TP + TN + FP + FN}
\end{equation}

\begin{equation}
\text{Precision} = \frac{TP}{TP + FP}
\end{equation}

\begin{equation}
\text{Recall} = \frac{TP}{TP + FN}
\end{equation}

\begin{equation}
F1 = 2 \times \frac{\text{Precision} \times \text{Recall}}{\text{Precision} + \text{Recall}}
\end{equation}

The results for each metric in the Table \ref{all_metrics} are demonstrated with the mean and standard deviation of the relevant metric across all folds. When we compare the deep CNN models in the reference article, it is observed that EfficientNet B1 achieves the best accuracy value of $0.932 \pm 0.086$. However, a subsequent analysis reveals that the 2D Self-ONN $(Q=2)$ model surpasses this accuracy, demonstrating superior diagnosis performance. Furthermore, the evaluation of metrics for the proposed 2D CNN $(Q=1)$ and the Self-ONN models reveals that the 2D Self-ONN $(Q=2)$ achieves the highest accuracy at $0.942 \pm 0.070$. Notably, each of the proposed Self-ONN models outperforms the 2D CNN $(Q=1)$ across all evaluated metrics, including accuracy. To illustrate the enhanced classification performance of the 2D Self-ONN $(Q=2)$ compared to the 2D CNN $(Q=1)$, their corresponding confusion matrices are provided in Table \ref{q2conf}.

\begin{table}[htbp]
\caption{Confusion matrices for 2D Self-ONN $(Q=2)$ and 2D CNN $(Q=1)$ in parentheses}
\begin{center}
\begin{tabular}{|c|c|c|c|}
\hline
\textbf{ }&\multicolumn{3}{|c|}{\textbf{Prediction}} \\
\hline
\textbf{Ground Truth} & \textbf{\textit{Healthy}}& \textbf{\textit{Misalignment}}& \textbf{\textit{Broken Rotor}} \\
\hline
\textbf{\textit{Healthy}}& $1936$ ($1741$)&$203$ ($186$)&$105$ ($317$)  \\ 
\hline
\textbf{\textit{Misalignment}}& $20$ ($102$)&$1779$ ($1697$)& $0$ ($0$) \\
\hline
\textbf{\textit{Broken Rotor}}&$0$ ($36$)&$0$ ($0$)&$1610$ ($1574$) \\
\hline
\end{tabular}
\label{q2conf}
\end{center}
\end{table}

Finally, Table \ref{param_inf} provides an overview of the number of trainable parameters and inference duration for each model in both the reference work and this study. Each model was executed 100 times on an NVIDIA GeForce GTX 1650 Ti GPU, and the average inference duration was computed and presented based on these runs. Table \ref{param_inf} shows that the EfficientNet B1 model achieves its accuracy with $6,517,027$ trainable parameters. In contrast, the 2D Self-ONN $(Q=2)$ model, which demonstrated the best performance among our proposals, achieves slightly higher classification accuracy with only $294,083$ parameters. This indicates that the 2D Self-ONN can achieve higher fault diagnosis accuracy while utilizing significantly fewer layers and trainable parameters than common CNN models used in this dataset. Furthermore, the proposed Self-ONNs exhibit considerably faster inference than all CNN models used in the reference study, emphasizing the computational efficiency of our approach.

\begin{table}[htbp]
\caption{Number of trainable parameters and average inference duration for each model}
\begin{center}
\begin{tabular}{|c|c|c|c|}
\hline
\multicolumn{2}{|c|}{\textbf{}} & \textbf{Trainable} & \textbf{Average Inference}  \\
\multicolumn{2}{|c|}{\textbf{Model}}& \textbf{Parameters} & \textbf{Duration (ms)}  \\ \hline
\multicolumn{2}{|c|}{PP-LCNet 50 \cite{bib13}}& $603,699$&$ 3.67$ \\ \hline
\multicolumn{2}{|c|}{PP-LCNet 75 \cite{bib13}}& $1,081,131$ & $3.71$ \\ \hline
\multicolumn{2}{|c|}{PP-LCNet 100 \cite{bib13}}& $1,676,643$ & $3.86$ \\ \hline
\multicolumn{2}{|c|}{SEMNASNet 50 \cite{bib13}}& $804,849$ & $7.84 $\\ \hline
\multicolumn{2}{|c|}{SEMNASNet 75 \cite{bib13}}& $1,635,121 $& $8.06 $\\ \hline
\multicolumn{2}{|c|}{SEMNASNet 100 \cite{bib13}}& $2,609,881$ &$ 8.08$ \\ \hline
\multicolumn{2}{|c|}{ResNet10t \cite{bib13}}& $4,924,027$ & $3.55$ \\ \hline
\multicolumn{2}{|c|}{ResNet18 \cite{bib13}} & $11,178,051$ &$ 6.43$ \\ \hline

\multicolumn{2}{|c|}{ResNet34 \cite{bib13}}& $21,286,211$ & $14.53$ \\ \hline

\multicolumn{2}{|c|}{EfficientNet B0 \cite{bib13}}& $4,011,391$ &$ 9.97$ \\ \hline

\multicolumn{2}{|c|}{EfficientNet B1 \cite{bib13}*}& \textbf{\boldmath{ $6,517,027$ }} & \textbf{\boldmath{ $14.97$ }} \\ \hline

\multicolumn{2}{|c|}{EfficientNet B2 \cite{bib13}}& $7,705,221$ & $16.15 $\\ \hline

\multicolumn{2}{|c|}{MixNet S \cite{bib13}}& $2,602,217$ & $12.05$ \\ \hline
 &\multicolumn{1}{l|} {$Q = 1$} & $293,027$ & $0.59$  \\ \cline{2-4} 
 & \multicolumn{1}{l|} {$Q = 2$**} & \textbf{\boldmath{ $294,083 $ }}& \textbf{\boldmath{ $ 0.61$}} \\ \cline{2-4} 
 2D Self-ONN   & \multicolumn{1}{l|}{$Q = 3$} & $295,139$ & $0.65$  \\ \cline{2-4} 
 & \multicolumn{1}{l|}{$Q = 4$} & $296,195$ & $0.69 $\\ \cline{2-4} 
 &\multicolumn{1}{l|} {$Q = 5$} & $297,251 $& $ 0.71$ \\ \hline
\multicolumn{4}{l}{* The best result from the reference study.} \\
\multicolumn{4}{l}{** The best result in all our proposed methods (2D Self-ONN).} 
\end{tabular}
\label{param_inf}
\end{center}
\end{table}

\section{Conclusion}
The thermal image-based fault diagnosis method offers a non-contact condition monitoring solution that can be seamlessly integrated into existing monitoring systems using infrared cameras. In this study, we employed 2D Self-ONNs to diagnose misaligment and broken rotor faults using a thermal image dataset collected from 3-phase squirrel cage induction motors. We evaluated the performance of the proposed 2D Self-ONN models with varying higher-order approximations, represented by different $Q$ values, against commonly used CNN models from \cite{bib13}. While EfficientNet B1 achieved the highest fault diagnosis accuracy of $0.932 \pm 0.086$ with $6,517,027$ parameters among all CNN models, the 2D Self-ONN $(Q=2)$ surpassed this with an accuracy of $0.942 \pm 0.070$, with only $294,083$ trainable parameters. Furthermore, all 2D Self-ONN models exhibit faster execution times on a GPU compared to the conventional deep CNN architectures. Self-ONNs achieve superior diagnostic performance with shallower architectures, highlighting their potential for efficient fault diagnosis. As a direction for future work, we intend to employ Self-ONNs as feature extractors within a domain-adversarial learning framework to enable fault diagnosis using thermal images across varying current loads.

\bibliography{references}

\begin{thebibliography}{10}
\providecommand{\url}[1]{#1}
\csname url@samestyle\endcsname
\providecommand{\newblock}{\relax}
\providecommand{\bibinfo}[2]{#2}
\providecommand{\BIBentrySTDinterwordspacing}{\spaceskip=0pt\relax}
\providecommand{\BIBentryALTinterwordstretchfactor}{4}
\providecommand{\BIBentryALTinterwordspacing}{\spaceskip=\fontdimen2\font plus
\BIBentryALTinterwordstretchfactor\fontdimen3\font minus \fontdimen4\font\relax}
\providecommand{\BIBforeignlanguage}[2]{{%
\expandafter\ifx\csname l@#1\endcsname\relax
\typeout{** WARNING: IEEEtran.bst: No hyphenation pattern has been}%
\typeout{** loaded for the language `#1'. Using the pattern for}%
\typeout{** the default language instead.}%
\else
\language=\csname l@#1\endcsname
\fi
#2}}
\providecommand{\BIBdecl}{\relax}
\BIBdecl

\bibitem{bib1_1}
R.~R. Kumar, M.~Andriollo, G.~Cirrincione, M.~Cirrincione, and A.~Tortella, ``A comprehensive review of conventional and intelligence-based approaches for the fault diagnosis and condition monitoring of induction motors,'' \emph{Energies}, vol.~15, no.~23, 2022.

\bibitem{bib1}
C.-C. Kuo, C.-H. Liu, H.-C. Chang, and K.-J. Lin, ``Implementation of a motor diagnosis system for rotor failure using genetic algorithm and fuzzy classification,'' \emph{Applied Sciences}, vol.~7, p.~31, 12 2016.

\bibitem{bib3}
A.~Khodja, G.~Noureddine, M.~Saadi, and N.~Boutasseta, ``Rolling element bearing fault diagnosis for rotating machinery using vibration spectrum imaging and convolutional neural networks,'' \emph{The International Journal of Advanced Manufacturing Technology}, vol. 106, pp. 1737--1751, 01 2020.

\bibitem{bib4}
C.~Celebioglu, S.~Kilickaya, and L.~Eren, ``Smartphone-based bearing fault diagnosis in rotating machinery using audio data and 1d convolutional neural networks,'' in \emph{Proceedings of the International Conference on Computer Systems and Technologies 2024}, 2024, pp. 149--154.

\bibitem{bib13}
M.~Piechocki, T.~Pajchrowski, M.~Kraft, M.~Wolkiewicz, and P.~Ewert, ``Unraveling induction motor state through thermal imaging and edge processing: A step towards explainable fault diagnosis,'' \emph{Eksploatacja i Niezawodność – Maintenance and Reliability}, vol.~25, no.~3, 2023.

\bibitem{bib5}
A.~Glowacz and Z.~Glowacz, ``Diagnosis of the three-phase induction motor using thermal imaging,'' \emph{Infrared Physics \& Technology}, vol.~81, pp. 7--16, 2017.

\bibitem{bib6}
A.~K. Al-Musawi, F.~Anayi, and M.~Packianather, ``Three-phase induction motor fault detection based on thermal image segmentation,'' \emph{Infrared Physics \& Technology}, vol. 104, p. 103140, 2020.

\bibitem{bib7}
A.~Glowacz, ``Thermographic fault diagnosis of electrical faults of commutator and induction motors,'' \emph{Engineering Applications of Artificial Intelligence}, vol. 121, p. 105962, 2023.

\bibitem{bib8}
Y.~Li, X.~Du, F.~Wan, X.~Wang, and H.~Yu, ``Rotating machinery fault diagnosis based on convolutional neural network and infrared thermal imaging,'' \emph{Chinese Journal of Aeronautics}, vol.~33, no.~2, pp. 427--438, 2020.

\bibitem{bib9}
L.~Xu, S.~S. Teoh, and H.~Ibrahim, ``A deep learning approach for electric motor fault diagnosis based on modified inceptionv3,'' \emph{Scientific Reports}, vol.~14, 05 2024.

\bibitem{bib10}
H.~Shao, W.~Li, M.~Xia, Y.~Zhang, C.~Shen, D.~Williams, A.~Kennedy, and C.~W. de~Silva, ``Fault diagnosis of a rotor-bearing system under variable rotating speeds using two-stage parameter transfer and infrared thermal images,'' \emph{IEEE Transactions on Instrumentation and Measurement}, vol.~70, pp. 1--11, 2021.

\bibitem{bib12}
A.~Choudhary, S.~Shimi, and A.~Akula, ``Bearing fault diagnosis of induction motor using thermal imaging,'' in \emph{2018 International Conference on Computing, Power and Communication Technologies (GUCON)}, 2018, pp. 950--955.

\bibitem{cnn_1}
T.~Ince, S.~Kiranyaz, L.~Eren, M.~Askar, and M.~Gabbouj, ``Real-time motor fault detection by 1-d convolutional neural networks,'' \emph{IEEE Transactions on Industrial Electronics}, vol.~63, no.~11, pp. 7067--7075, 2016.

\bibitem{cnn_2}
C.-C. Chen, Z.~Liu, G.~Yang, C.-C. Wu, and Q.~Ye, ``An improved fault diagnosis using 1d-convolutional neural network model,'' \emph{Electronics}, vol.~10, no.~1, 2021.

\bibitem{kiranyaz2021self}
S.~Kiranyaz, J.~Malik, H.~B. Abdallah, T.~Ince, A.~Iosifidis, and M.~Gabbouj, ``Self-organized operational neural networks with generative neurons,'' \emph{Neural Networks}, vol. 140, pp. 294--308, 2021.

\bibitem{rotor_onn}
L.~Eren, O.~C. Devecioglu, T.~Ince, and M.~Askar, ``Improved detection of broken rotor bars by 1-d self-onns,'' in \emph{IECON 2022 – 48th Annual Conference of the IEEE Industrial Electronics Society}, 2022, pp. 1--5.

\bibitem{bib2}
T.~Ince, S.~Kilickaya, L.~Eren, O.~C. Devecioglu, S.~Kiranyaz, and M.~Gabbouj, ``Improved domain adaptation approach for bearing fault diagnosis,'' in \emph{IECON 2022 – 48th Annual Conference of the IEEE Industrial Electronics Society}, 2022, pp. 1--6.

\bibitem{malik2021self}
J.~Malik, S.~Kiranyaz, and M.~Gabbouj, ``Self-organized operational neural networks for severe image restoration problems,'' \emph{Neural Networks}, vol. 135, pp. 201--211, 2021.

\bibitem{kiranyaz2020operational}
S.~Kiranyaz, T.~Ince, A.~Iosifidis, and M.~Gabbouj, ``Operational neural networks,'' \emph{Neural Computing and Applications}, vol.~32, no.~11, pp. 6645--6668, 2020.

\bibitem{cam_wic}
\BIBentryALTinterwordspacing
Workswell, ``Workswell infrared camera (wic),'' 2024. [Online]. Available: \url{https://workswell.eu/thermal-camera-for-production-control/}
\BIBentrySTDinterwordspacing

\bibitem{pplcnet}
\BIBentryALTinterwordspacing
C.~Cui, T.~Gao, S.~Wei, Y.~Du, R.~Guo, S.~Dong, B.~Lu, Y.~Zhou, X.~Lv, Q.~Liu, X.~Hu, D.~Yu, and Y.~Ma, ``Pp-lcnet: A lightweight cpu convolutional neural network,'' 2021. [Online]. Available: \url{https://arxiv.org/abs/2109.15099}
\BIBentrySTDinterwordspacing

\bibitem{resnet10}
J.~Gong, W.~Liu, M.~Pei, C.~Wu, and L.~Guo, ``Resnet10: A lightweight residual network for remote sensing image classification,'' in \emph{2022 14th International Conference on Measuring Technology and Mechatronics Automation (ICMTMA)}, 2022, pp. 975--978.

\bibitem{MNASNet}
M.~Tan, B.~Chen, R.~Pang, V.~Vasudevan, M.~Sandler, A.~Howard, and Q.~V. Le, ``Mnasnet: Platform-aware neural architecture search for mobile,'' in \emph{2019 IEEE/CVF Conference on Computer Vision and Pattern Recognition (CVPR)}, 2019, pp. 2815--2823.

\bibitem{efficientnet}
M.~Tan and Q.~V. Le, ``Efficientnet: Rethinking model scaling for convolutional neural networks,'' \emph{arXiv}, 2020.

\bibitem{adamW}
\BIBentryALTinterwordspacing
I.~Loshchilov and F.~Hutter, ``Decoupled weight decay regularization,'' 2019. [Online]. Available: \url{https://arxiv.org/abs/1711.05101}
\BIBentrySTDinterwordspacing

\bibitem{adam}
\BIBentryALTinterwordspacing
D.~P. Kingma and J.~Ba, ``Adam: A method for stochastic optimization,'' 2017. [Online]. Available: \url{https://arxiv.org/abs/1412.6980}
\BIBentrySTDinterwordspacing

\end{thebibliography}
\bibliographystyle{IEEEtran}

\end{document}